\pdfoutput=1

\documentclass[11pt]{article}

\usepackage[final]{acl}

\usepackage{times}
\usepackage{latexsym}

\usepackage[T1]{fontenc}

\usepackage[utf8]{inputenc}

\usepackage{microtype}

\usepackage{inconsolata}

\usepackage{graphicx}
\usepackage{amsmath}
\usepackage{booktabs}
\usepackage{multirow}
\usepackage{tcolorbox}
\usepackage{listings}
\usepackage{algorithm}
\usepackage{algorithmic}

\title{EcoLANG: Efficient and Effective Agent Communication Language Induction for Social Simulation}

 \author{
    Xinyi Mou$^1$,
    Chen Qian$^2$,
    Wei Liu$^3$,
    Xuanjing Huang$^{1}$,
    Zhongyu Wei$^{1,4}$\Thanks{ Corresponding author.}\\
    \textsuperscript{\rm 1}Fudan University, \textsuperscript{\rm 2}Shanghai Jiao Tong University\\
    \textsuperscript{\rm 3}King's College London, \textsuperscript{\rm 4}Shanghai Innovation Institute\\
    \texttt{\{xymou20,xjhuang,zywei\}@fudan.edu.cn}\\
    \texttt{qianc@sjtu.edu.cn}, \texttt{wei.4.liu@kcl.ac.uk}
    \ \ \ 
    }

\begin{document}
\maketitle
\begin{abstract}
Large language models (LLMs) have demonstrated an impressive ability to role-play humans and replicate complex social dynamics. While large-scale social simulations are gaining increasing attention, they still face significant challenges, particularly regarding high time and computation costs. Existing solutions, such as distributed mechanisms or hybrid agent-based model (ABM) integrations, either fail to address inference costs or compromise accuracy and generalizability. To this end, we propose \textbf{EcoLANG}: \underline{E}fficient and Effective Agent \underline{Co}mmunication \underline{Lang}uage Induction for Social Simulation. EcoLANG operates in two stages: (1) language evolution, where we filter synonymous words and optimize sentence-level rules through natural selection, and (2) language utilization, where agents in social simulations communicate using the evolved language. Experimental results demonstrate that EcoLANG reduces token consumption by over 20\%, enhancing efficiency without sacrificing simulation accuracy~\footnote{Available at \url{https://github.com/xymou/EcoLANG} .}.

\end{abstract}

\section{Introduction}
Social simulation, which explores the dynamics and emergent behaviors of social systems by modeling interactions among individuals, has become a powerful approach for understanding complex societal phenomena~\cite{squazzoni2014social,mou2024individual}. Recent advancements in large language models (LLMs) have further expanded the potential of this field, enabling agents to simulate human behavior at various levels, from mimicking well-known individuals~\cite{argyle2023out,park2024generative} and reconstructing specific scenarios for task-solving~\cite{hong2023metagpt,qian2024chatdev} to modeling large-scale social dynamics~\cite{mou-etal-2024-unveiling,li2024econagent,zhang2024generative}. Among these efforts, large-scale social simulation focuses on the emergence of collective behaviors. Rather than perfectly replicating each individual's wording, researchers aim to capture macro-level phenomena such as opinion divergence and the evolution of public discourse, insights that are crucial for applications in social governance, information dissemination, and crisis management.

\begin{figure}[!t]
    \centering
    \includegraphics[width=\linewidth]{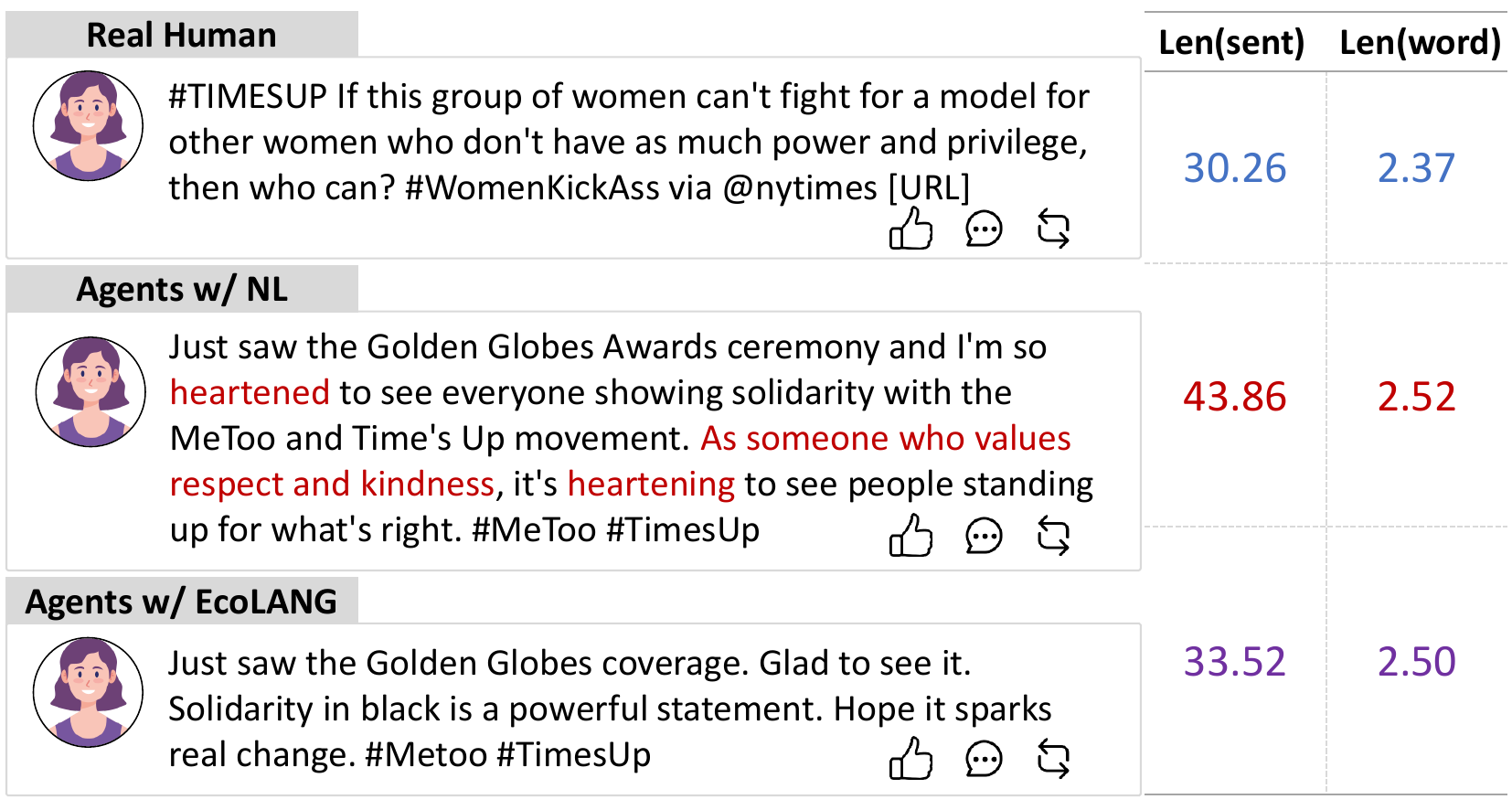}
    \caption{Responses generated by humans and agents when discussing the MeToo movement. There is information redundancy in the vanilla setting, i.e., Agents w/ natural language (NL), such as long but unnecessary sentences and advanced but uncommon words (left). The average sentence length (\textit{Len(sent)}) and word length (\textit{Len(word})) of responses further highlight this issue in current agent communication with NL, and this can be mitigated by our proposed EcoLANG (right).}
    \label{fig:intro}
\end{figure}

However, conducting large-scale social simulations with LLMs remains highly challenging. Simulating thousands of agents across millions of interactions incurs substantial time and computational costs, posing a major barrier to their practical deployment in real-world scenarios~\cite{gao2024large}. Current efforts to mitigate this challenge generally fall into two categories: (1) Some works enhance efficiency by deploying open-source models in distributed mechanisms~\cite{pan2024very,yang2024oasis}, thereby avoiding the high costs associated with closed-sourced models. However, these methods \textbf{do not fundamentally resolve the issue of inference overhead}. (2) Other works increase scalability by simplifying the modeling of most agents, either through integration with agent-based modeling (ABM) frameworks~\cite{chopra2024limits,mou-etal-2024-unveiling} or by reusing certain strategies~\cite{yu2024affordable}. While these methods significantly improve simulation feasibility, they \textbf{often compromise accuracy and may lack generalizability across diverse scenarios}.

To better understand the inefficiencies in current simulations, we revisit the characteristics of real-world social communication and those between agents. As shown in Figure~\ref{fig:intro}, current LLM-driven multi-agent communication exhibits obvious \textbf{communication redundancy}. Agents tend to generate verbose and overly elaborate responses, whereas humans naturally pursue efficient communication with minimal effort~\cite{zipf2016human}, typically using simple, familiar words to reduce cognitive load and concise expressions to save time. This observation highlights that current agent language not only increases inference costs but also diverges from natural human communication patterns. Motivated by this, we explore a novel direction: enhancing the efficiency of social simulations by developing a simplified agent language for more concise and effective communication.

In this paper, we introduce \textbf{EcoLANG}: \underline{E}fficient and Effective Agent \underline{Co}mmunication \underline{Lang}uage Induction for Social Simulation. EcoLANG operates in two stages: \textbf{language evolution} and \textbf{language utilization}. In the evolution stage, inspired by the principle of least effort~\cite{zipf2016human}, we first construct a compact vocabulary by filtering synonymous words based on frequency and length, thereby reducing the vocabulary size of the underlying LLMs. We then adopt a natural selection paradigm, prompting agents to communicate using different rule sets in dialogue-intensive scenarios and iteratively optimizing these rules to evolve efficient, sentence-level communication strategies. In the utilization stage, we enforce the use of the induced language in large-scale social simulations by modifying the inference model’s vocabulary and integrating rules. Since this language is induced through general communication and does not rely on any task-specific architecture, it is framework-agnostic, allowing it to adapt to various scenarios.

We conduct extensive experiments on the open-source Llama-3.1-8B-Instruct~\cite{dubey2024llama}. We induce the language on twitter corpus and the synthetic-persona-chat dataset~\cite{jandaghi2023faithful}, and validate the effectiveness of this language in social simulations on PHEME~\cite{zubiaga2016pheme} and HiSim~\cite{mou-etal-2024-unveiling}. Experiment results show that EcoLANG can significantly reduce token consumption and improve the efficiency of social simulations without compromising simulation accuracy, demonstrating advantages over baselines such as structured languages and traditional agent communication languages. Overall, our contributions can be summarized as follows:
\begin{itemize}
    \item We introduce EcoLANG, a two-stage paradigm consisting of language evolution and language utilization. EcoLANG can induce efficient and effective language for LLM-driven social simulations.
    \item We derive a compact, generalizable agent language using EcoLANG, induced from the Twitter corpus and synthetic-persona-chat dataset, and show its generalizability across diverse downstream social simulation tasks.
    \item We perform extensive experiments across different scenarios, demonstrating that EcoLANG reduces inference costs while preserving simulation accuracy across different levels of granularity.
\end{itemize}

\section{Related Work}
\begin{figure*}[!t]
    \centering
    \includegraphics[width=\linewidth]{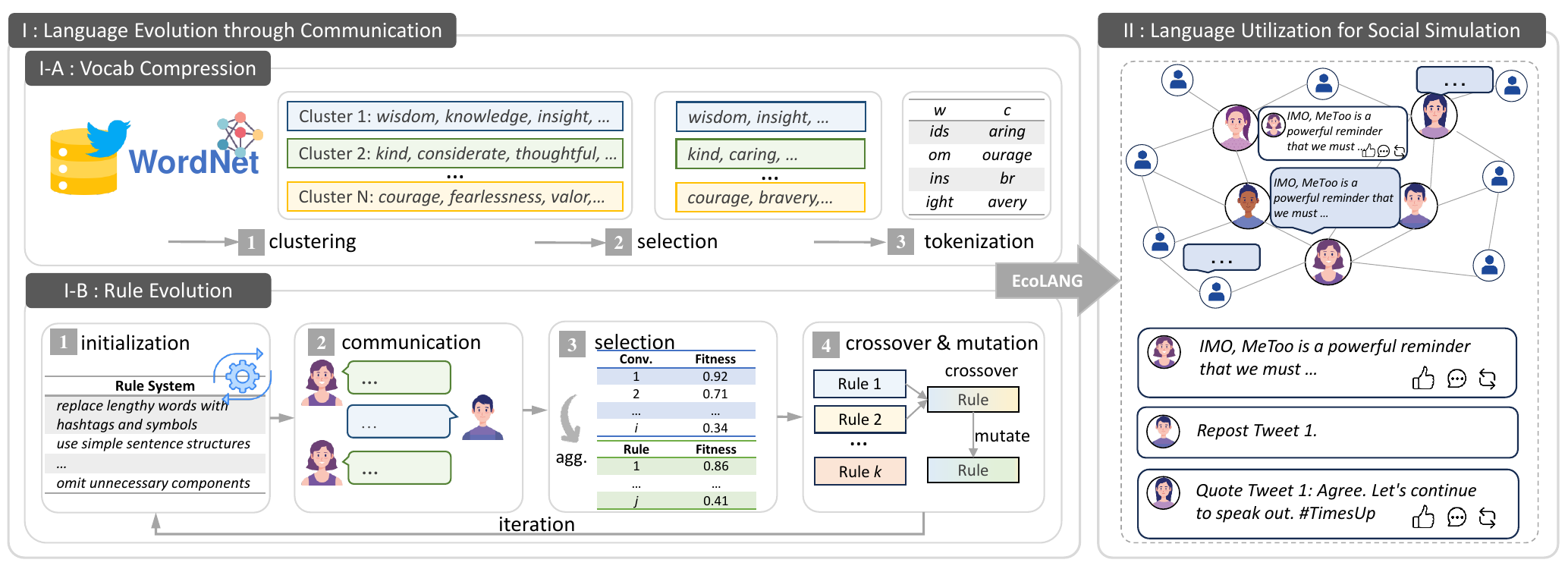}
    \caption{Overview of the EcoLANG framework. We get the language through vocabulary compression and rule evolution in dialogue-intensive scenarios. Then, we enable agents to use this language in social simulations.}
    \label{fig:fm}
\end{figure*}

\subsection{LLM-driven Social Simulation}
Recently, LLMs have been used to construct agents to empower social simulations, aiming to reveal and explain emergent behaviors and the outcomes of interactions among numerous agents~\cite{mou2024individual}. In such simulations, each agent role-plays a person in society and participates in social interactions, with the goal of modeling complex phenomena such as opinion dynamics~\cite{chuang2024simulating}, epidemic modeling~\cite{williams2023epidemic}, and macroeconomic activities~\cite{li2024econagent}. Initial researches construct virtual spaces supporting such simulations~\cite{park2022social,park2023generative}. Further studies focus on alignment on specific scenarios, validating whether real-world behaviors and phenomena can be replicated by such simulations~\cite{gao2023s,zhang2025socioverse}. Although LLMs show potential in mimicking human, their integration into large-scale simulations remains challenging. Some work has improved simulation efficiency by deploying open-source model-driven agents through distributed mechanisms~\cite{pan2024very,yang2024oasis}, but it has not addressed the fundamental issue of computational costs and communication efficiency. 
Other work seeks to combine with agent-based models, simplifying the modeling of certain agents, which may sacrifice some simulation effectiveness~\cite{mou-etal-2024-unveiling,chopra2024limits}.

\subsection{Multi-Agent Communication}
Before the rise of LLMs, some studies focused on how multi-agent systems could use language to cooperate in completing tasks or solving problems~\cite{havrylov2017emergence,lazaridou2020emergent,lazaridou2020multi}, typically developing effective communication protocols with task success as a training signal. In current LLM-driven multi-agent systems, communication is mainly conducted through natural language. Some research has highlighted the redundancy in communication, leading to suggestions that agents autonomously choose structured languages like JSON for communication~\cite{chen-etal-2024-beyond-natural,marro2024scalable} or further fine-tune models to improve this communication~\cite{chen2024optima}. Meanwhile, other studies have approached communication optimization from the perspective of its structure, aiming to enhance efficiency by pruning the spatial-temporal message  graph~\cite{zhang2024cut}. However, most existing work focuses on task-solving rather than social simulation, which more urgently needs to address the challenges of large-scale simulation.

\section{Methodology}
\subsection{Overview}
To address the challenges of high cost and inefficiency in social simulations, particularly the generation of unnecessary content, we propose a two-stage paradigm EcoLANG. First, we reconstruct the vocabulary based on the principle of least effort (Sec~\ref{sec:vocab}) and evolve a set of rules through natural selection (Sec~\ref{sec:rule}). Once the new language is induced, we apply it to social simulations by encouraging agents to adopt this language for communication (Sec~\ref{sec:use}). The overall process is shown in Figure ~\ref{fig:fm}.

\subsection{Vocabulary Compression}\label{sec:vocab}
The development of a new language begins with defining its basic elements, i.e., the vocabulary. Since LLMs are primarily trained on natural languages, we don't aim to introduce an entirely novel symbolic language or completely replace the existing vocabulary. Instead, we focus on compressing the vocabulary. In natural languages, many words have similar meanings, and low-frequency words are often substituted with more common synonyms~\cite{mohammad2020wordwars}. This reflects the principle of least effort~\cite{zipf2016human}, which suggests that people tend to communicate in the most efficient way possible. Thus, we compress the vocabulary of LLMs, as outlined in the following steps and illustrated in part I-A of Figure~\ref{fig:fm}.

\paragraph{Semantic Clustering}
To ensure that the induced language retains sufficient expressive power, the new vocabulary must include words that span a broad range of semantics. To this end, we begin by clustering all words in the corpus based on their semantic similarity, followed by selective filtering within each semantic group. Rather than clustering from scratch, we leverage existing synsets from WordNet~\cite{miller1995wordnet} and assign each word to the most semantically relevant synset using embedding similarity. For each word $w_i$, we compute the similarity between its embedding $\mathbf{e}{w_i}$ and the centroid embedding of each synset $\mathbf{e}{S_j}$, and assign $w_i$ to the synset with the highest similarity:
\begin{align}
    S(w_i)=\arg \max _j\left(\operatorname{sim}\left(\mathbf{e}_{w_i}, \mathbf{e}_{S_j}\right)\right), \label{eq1}
\end{align}
This synset-based clustering approach improves controllability and reduces noise compared to unsupervised clustering methods.

\paragraph{Intra-Cluster Selection}
Within each cluster, we further filter words by assigning a score that balances two key factors: word frequency and word length. Frequent words are generally more effective at conveying intent and are better supported by the LLM due to their higher training occurrence. Meanwhile, shorter words are preferred to reduce the length of generated outputs. To integrate these considerations, we define the following scoring function:
\begin{align}
    R(w_i)=\lambda_{freq}F(w_i)+\lambda_{token}(1-L(w_i)),  \label{eq2}
\end{align}
where $F(w_i)$ and $L(w_i)$ denote the percentile scores of the word's frequency and token length respectively. $\lambda_{freq}$ and $\lambda_{token}$ are hyperparameters controlling factors' relative importance. Given these scores, we retain the top words within each cluster according to a predefined retention ratio $r_w$. 

\paragraph{Tokenization}
While humans communicate using words as basic units, LLMs process text in units of tokens. Therefore, after selecting the target words to retain, we tokenize them to identify the corresponding tokens to be preserved. Although these tokens may form additional words beyond our initial selection, the overall vocabulary size of the LLMs is still effectively reduced. Moreover, we ensure that special tokens, which are essential for the model's correct generation, are preserved throughout this process.

\subsection{Language Rule Evolution}\label{sec:rule}
Once vocabulary, the fundamental elements of a language is established, the next core component is the organization of these elements: the rule system. Previous research in linguistics ~\cite{pinker1990natural,nowak1999evolution} suggest that grammar functions as a simplified set of rules evolved through natural selection to minimize communication errors. Inspired by this, we design our rule system following the principles of evolutionary algorithms (EAs). The task is formulated as identifying rules or prompts that allow agents to communicate both effectively and efficiently. The overall process is outlined in part I-B of Figure~\ref{fig:fm}.

\paragraph{Initialization}
The evolutionary process begins with an initial population of $N$ solutions, represented as rule prompts $\mathcal{P}=\left\{p_1, p_2, \ldots, p_N\right\}$, which are iteratively refined to generate improved solutions. To initialize the rule system, we employ a combination of manually crafted prompts and those generated by LLMs, to leverage the wisdom of humans and LLMs~\cite{guoconnecting}. These prompts are designed to guide agents toward concise expression. More details can be found in Appendix~\ref{app:evo_init}.

\paragraph{Communication}
Language emerges and evolves through communication within social interactions. To observe how individuals using specific rules communicate, we simulate dialogues between LLM-driven agents. Given a set of dialogue scenarios $\mathcal{D}$ between two agents, for each scenario $d_i \in \mathcal{D}$, we generate $M$ dialogue trajectories $\left\{\tau_i^j\right\}_{j=1}^M$, each with a randomly selected rule from $\mathcal{P}$ appended to the original prompt to the agents.

\paragraph{Selection}
To select high-quality rules, we evaluate the dialogue trajectories using a fitness function. A good language should be both effective and efficient. We measure \textit{efficiency} by the number of tokens used. For effectiveness, while prior work in multi-agent collaboration often relies on task success rates~\cite{lazaridou2020multi}, social simulation lacks explicit tasks. Instead, we propose to adopt \textit{alignment}—how well an agent embodies its assigned persona, as the key indicator of effectiveness. Furthermore, \textit{expressiveness}~\cite{galke2022emergent} is crucial to prevent the language from becoming overly abstract and to maintain fluency. Integrating these considerations, we define the fitness of a dialogue trajectory $\tau_i^j$ as follows:
\begin{align}
    \begin{split}
F(\tau_i^j)&=\lambda_{align}Align(\tau_i^j)+\lambda_{eff}Eff(\tau_i^j)\\
&\quad +\lambda_{exp}Exp(\tau_i^j), \label{eq3}
    \end{split}
\end{align}
where the alignment score $Align(\tau_i^j)$ and the expressiveness score $Exp(\tau_i^j)$ are given by external judge LLMs, and $Eff(\tau_i^j)$ is the normalized token count $\frac{\# \operatorname{Tokens}\left(\tau_i^j\right)}{\max _k\left(\left\{\# \operatorname{Tokens}\left(\tau_i^k\right)\right\}_k\right)}$. $\lambda_{align}$, $\lambda_{eff}$ and $\lambda_{exp}$ are hyperparameters controlling factors' relative importance. Finally, we aggregate and average the fitness scores of all trajectories associated with each rule to derive that rule’s overall fitness.

\paragraph{Crossover and Mutation}
To encourage rule diversity, we apply crossover and mutation operations to the high-quality rules selected through the selection process. Specifically, parent rules are sampled from the population with probabilities proportional to their fitness scores. We then prompt LLMs to generate new rule candidates, following the strategy introduced by~\citeauthor{guoconnecting}.

\paragraph{Update and Iteration}
In each iteration, we use the elitism strategy of genetic algorithm to update the population. The top $N/2$ rules with the highest fitness scores are retained from the current population, while the remaining $N/2$ are generated through crossover and mutation operations. This ensures that the population size remains fixed at $N$. In applications, the highest-scoring rule is selected as the final rule of the newly evolved language. The complete procedure is summarized in Algorithm~\ref{alg:evo}.

\subsection{Language Utilization in Social Simulation}\label{sec:use}

Once the vocabulary and rule system of the new language are established, we enable agents to communicate in this language by modifying the decoding range of the LLMs and incorporating the rules into their contextual prompts. Although the language can be evolved within the same social simulation scenario where it will be used, we adopt a \textbf{transfer setting}, in which the language is evolved on \textbf{general multi-turn dialogue} data and applied to \textbf{downstream social simulation} tasks. This is motivated by two key considerations: (1) large-scale social simulation data is often sparse, whereas general dialogue data offers richer and more intensive communication, facilitating more efficient language evolution; (2) new languages emerge naturally from everyday conversations, making general dialogues a more task-agnostic and robust foundation for language development.

\begin{table*}[!t]
\resizebox{\textwidth}{!}{
\begin{tabular}{@{}c|cccccc|ccccccc@{}}
\toprule
\multirow{2}{*}{Method} & \multicolumn{6}{c|}{PHEME}                                                                                         & \multicolumn{7}{c}{HiSim}                                                                                                                          \\
                        & stance↑  & belief↑   & belief\_JS↓  & $token_r$↓ & $token_p$↓   & $token_c$↓    & stance↑& content↑   & $\Delta{bias}$ ↓   & $\Delta{div}$ ↓    & $token_r$↓  & $token_p$↓    & $token_c$↓    \\ \midrule
Base                   & 66.21 & 42.44 & 0.137 & 2.61K & 84.43K & 8.44K & 70.30 & 30.23 & 0.093 & 0.027& 13.02K & 1.92M  & 283.79K \\
Summary                 & 66.07 & 41.55 & 0.133 & 2.41K & 84.27K & 8.01K & \textbf{70.95} & 32.31 & 0.089 & \underline{0.023} & 10.62K & 1.90M & 269.73K \\
AutoForm                & 63.72 & 40.50 & 0.136 & \underline{2.00K} & 85.02K & \underline{7.69K} & 69.92 & 32.04 & \textbf{0.082} & 0.029 & 10.66K & 1.89M  & 252.09K \\
KQML                   & 57.66 & 42.09 & 0.130 & 3.01K & 91.10K & 9.18K & 70.16 & \underline{32.47} & 0.093 & 0.032 & 12.06K & 1.96M  & 279.17K \\ \midrule
Vocab                   & 65.73 & 44.65 & 0.131 & 2.67K  & 84.78K  & 8.70K & 70.34 & 30.48 & 0.086 & \underline{0.023} & 11.37K & 1.91M & 286.41K \\
Rule                    & \textbf{66.86} & \underline{45.14} & \underline{0.128} & \textbf{1.98K }& \textbf{82.08K} & \textbf{7.52K }& \underline{70.63} & 32.25 & 0.091 & 0.027 & \underline{9.07K}  & 1.84M  & 242.43K\\
EcoLANG                 & \underline{66.34} & \textbf{45.50} & \textbf{0.128} & 2.08K & \underline{82.26K} & 7.70K & 70.60 & \textbf{32.57} & \underline{0.083} & \textbf{0.020} & 9.80K  & \underline{1.83M}  & \underline{236.83K} \\ \bottomrule
\end{tabular}
}
\caption{Experimental results of different methods. The average results of 3 runs are reported. We report the best performance
in \textbf{bold} format and the second best in \underline{underlined} format.}
\label{tab:exp_main}
\end{table*}
\section{Experiment Settings}
As mentioned before, we evolve and utilize language in different scenarios. We filter the vocabulary using a Twitter corpus and acquire rules from dialogue-intensive scenarios. The language is then applied to social simulation scenarios, namely PHEME~\cite{zubiaga2016pheme} and the Metoo and Roe events from HiSim~\cite{mou-etal-2024-unveiling}. PHEME simulates the propagation and discussion of potential rumors, while HiSim models the evolution of opinion dynamics following the release of triggering news related to social movements.

\subsection{Language Evolution Settings}
\paragraph{Twitter Corpus for Vocabulary Compression}
As our vocabulary filtering in Sec.\ref{sec:vocab} partially relies on word frequency, we require a corpus to compute word statistics. While ideally we would analyze all tweets, this is impractical. Instead, we collect and analyze tweets related to the topics of our social simulation scenarios. Therefore, we have chosen to analyze and gather statistics from existing tweets relevant to the topics of our social simulation scenarios. Specifically, for PHEME, which models rumor propagation, we use tweets from Twitter15\cite{liu2015real} and Twitter16~\cite{ma2016detecting}. For HiSim, we use tweets posted prior to the simulated events~\cite{metoodata,chang2023roeoverturned,mou-etal-2024-unveiling}.

\paragraph{Scenarios for Communication in Rule Evolution}
For rule evolution, we use the synthetic-persona-chat dataset~\cite{jandaghi2023faithful} to generate dialogues between agents adhering to specific language rules. This dataset provides a collection of dialogues between two users with diverse personalities, along with their corresponding personality descriptions. We provide these profiles to LLMs and instruct them to role-play the corresponding individuals in conversation, thereby obtain dialogue trajectories for further selection.

\paragraph{Implementation Details}
The agents are powered by Llama-3.1-8B-Instruct~\cite{dubey2024llama}. For vocabulary compression, we set the hyperparameters $\lambda_{freq}=1$, $\lambda_{token}=1$. The reservation ratio $r_w$ for each semantic cluster is 0.6 for PHEME and 0.2 for HiSim, yielding vocabulary sizes of 32.6K (25.4\% of Llama-3.1's vocabulary) and 48.2K (37.5\% of Llama-3.1's vocabulary), respectively. For rule evolution, we initialize the population with $N=10$ rules. The development set of the synthetic-persona-chat dataset, which contains 1,000 dialogue scenarios, is used to simulate communication during the evolution process. For selection, GPT-4o~\cite{achiam2023gpt} serves as the judge to evaluate alignment and expressiveness based on reference dialogues. The weight hyperparameters are  set to $\lambda_{align}=1$, $\lambda_{eff}=0.6$ and $\lambda_{exp}=0.6$. In each iteration, We retain the top-5 parent rules and generate 5 new rules via crossover and mutation, with parents sampled in proportion to their fitness scores. The evolution process is run for 5 iterations. Additional details are in Appendix~\ref{app:evo}.

\subsection{Language Utilization Settings}
\paragraph{Datasets}
From PHEME~\cite{zubiaga2016pheme}, we collect 196 real-world instances, each involving 2 to 31 users discussing a source tweet, to examine whether agents can mimic user responses towards rumors. From HiSim~\cite{mou-etal-2024-unveiling}, we use the second events of \#Metoo and \#Roe movements, each comprising 1,000 users discussing news related to the events over time, to examine the opinion dynamics in socially interactive settings.

\paragraph{Metrics}
For PHEME, we focus on content-related metrics. We measure consistency between each agent's initial stance on the source tweet and real users' stances, categorized into four types as in~\cite{derczynski2017semeval} and annotated by GPT-4o-mini. Following~\cite{liufps}, we also label each agent's final belief as \textit{belief}, \textit{disbelief}, or \textit{unknown} using GPT-4o-mini, and compute belief consistency and JS divergence~\cite{lin1991divergence} between the simulated and real-world belief distributions. More details about the experimental setup can be found in Appendix~\ref{app:pheme_smiu}.

For HiSim, we report stance and content consistency between agents’ initial responses and those of real users, again using GPT-4o-mini for labeling. We also report $\Delta{bias}$ and $\Delta{div.}$ to measure the difference in average opinion bias and diversity between simulated and real user groups over time. More details about the experimental setup can be found in Appendix~\ref{app:hisim_simu}.

For both datasets, we evaluate communication efficiency by reporting the average number of tokens in generated tweet responses per scenario (\# $tokens_r$), as well as the total token consumption per scenario, which includes both prompt tokens (\# $tokens_p$) and completion tokens (\# $tokens_c$).

\paragraph{Baselines}
We compare our method against the following communication strategies: (1) \textit{Base}: standard simulation without any additional rule prompts; (2) \textit{Summary}: agents are prompted to summarize their opinions when responding, as concise expression resembles a summarization task; (3) \textit{AutoForm}~\cite{chen-etal-2024-beyond-natural}: agents are prompted to automatically choose a structured format to respond, such as JSON and logical expression; (4) \textit{KQML}~\cite{finin1994kqml}: agents are prompted to use a traditional agent communication language KQML; (5) \textit{Vocab}: a variant of our method that only compresses the vocabulary of the LLMs; (6) \textit{Rule}: a variant of our method that only applies the evolved communication rules. Besides, we conduct additional experiments integrating these communication optimization methods with an ABM-based scalable simulation framework AgentTorch~\cite{chopra2024limits}, which clusters all agents into a small number of archetypes, simulates the actions of these archetypes, and then maps their responses to the corresponding agents.

\paragraph{Implementation Details}
Agents are powered by Llama-3.1-8B-Instruct~\cite{dubey2024llama}, and all simulations are conducted within the OASIS framework~\cite{yang2024oasis}. Each simulation is run three times, and we report the average results. We use GPT-4o-mini to label the stance, belief and content of responses and apply Textblob to calculate the opinion score. Further implementation details can be found in Appendix~\ref{app:simu}.

\begin{table*}[!t]
\centering
\resizebox{0.8\textwidth}{!}{
\begin{tabular}{@{}c|ccccc|ccccc@{}}
\toprule
\multicolumn{1}{l|}{\multirow{2}{*}{Method}} & \multicolumn{5}{c|}{PHEME}                                              & \multicolumn{5}{c}{HiSim}                                             \\
\multicolumn{1}{l|}{}                                 & CosSim↑        & Jaccard↑       & word\_JS↓            & $\Delta_{l_s}$↓  & $\Delta_{l_w}$↓ & CosSim↑        & Jaccard↑       & word\_JS↓            & $\Delta_{l_s}$↓ & $\Delta_{l_w}$↓ \\ \midrule
Base                                                  & \underline{0.662}          & 0.037          & 0.307          & 31.63          & 0.12          & \underline{0.782}          & \textbf{0.067} & 0.321          & 13.60         & 0.15          \\
Summary                                               & 0.651          & 0.039          & 0.306          & 29.94          & 0.10          & 0.769          & 0.062          & 0.320          & 9.98          & 0.14          \\
AutoForm                                              & 0.662          & 0.039          & 0.306          & 23.63          & 0.18          & 0.763          & 0.061          & 0.324          & 5.38          & 0.15          \\
KQML                                                  & 0.653          & 0.035          & 0.307          & 32.57          & 0.16          & 0.757          & 0.061          & 0.321          & 8.22          & \textbf{0.10} \\ \midrule
Vocab                                                 & \textbf{0.671} & \underline{0.039}          & 0.305          & 30.79          & \underline{0.06}          & \textbf{0.789} & 0.064          & 0.320          & 10.20         & \underline{0.12}          \\
Rule                                                  & 0.661          & 0.039          & \underline{0.299}          & \underline{22.43}          & 0.09          & 0.774          & 0.062          & \underline{0.319}          & \underline{4.25}          & 0.15          \\
EcoLANG                                               & 0.661          & \textbf{0.040} & \textbf{0.298} & \textbf{22.33} & \textbf{0.05} & 0.775          & \underline{0.065}          & \textbf{0.309} & \textbf{3.26} & 0.13          \\ \bottomrule
\end{tabular}
}
\caption{Comparison of semantic similarity and length between agent responses and real user responses. We report the best performance
in \textbf{bold} format and the second best in \underline{underlined} format.}
\label{tab:exp_fine}
\end{table*}
\section{Experiment Results}
\subsection{Overall Performance}
Table~\ref{tab:exp_main} presents the overall results, from which we make the following observations.

\noindent\textbf{(1) Can reducing communication redundancy improve simulation efficiency?} All simplified communication methods significantly reduce token usage compared to \textit{Base}. This improvement in efficiency is not only reflected in $token_r$ but also cumulatively transmitted to $token_p$ and $token_c$ as the generated content contributes to subsequent context via memory mechanisms and social interactoions. Among the approaches, our proposed \textit{Rule} and \textit{EcoLANG} are the most prominent, reducing generated tokens by over 20\%. Appendix~\ref{app:agenttorch} further shows that combining these methods with AgentTorch, an approach that alters the simulation paradigm, can further boosts efficiency, with our method achieving the best. However, AgentTorch comes at the cost of reduced diversity and accuracy of agent-generated content, suggesting that such paradigms should be used with caution.

\noindent\textbf{(2) Does simplifying communication compromise the simulation effectiveness?}
Some baselines such as \textit{AutoForm} and \textit{KQML}, despite enhancing efficiency, reduced the accuracy of the simulation. This may suggest that while these structured languages can improve the efficiency and effectiveness of task-solving, they might not be suitable for social simulation, as humans generally communicate using natural language. By contrast, benefiting from the considerations of both efficiency and alignment during the process of language evolution, our method is able to enhance efficiency while maintaining leading simulation accuracy.

\noindent\textbf{(3) Does vocabulary compression enhance performance or efficiency?}
Vocabulary compression does not significantly affect token usage, as changes in individual word lengths do not substantially alter overall sentence lengths. However, simulations still achieve comparable or even better performance after compression, e.g., in HiSim, suggesting that standard LLM vocabularies may be redundant for such social simulations. Theoretically, removing these tokens from LLM's vocabulary could enhance the model's inference efficiency and reduce GPU memory usage.

\begin{figure}[!t]
    \centering
    \includegraphics[width=\linewidth]{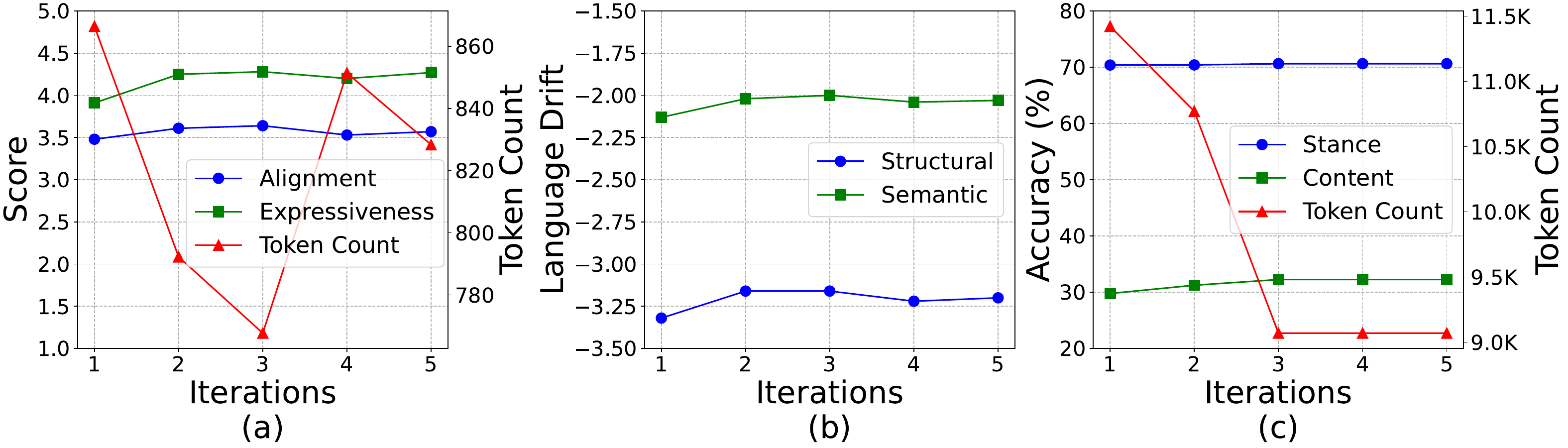}
    \caption{(a) Average fitness score change and (b) language drift change on synthetic-persona-chat simulated dialogues across iterations; (c) Performance and token consumption in HiSim using the best language rules acquired across iterations.}
    \label{fig:iter}
\end{figure}


\begin{table}[!t]
\resizebox{0.48\textwidth}{!}{
\begin{tabular}{@{}cccc|cccc@{}}
\toprule
\multicolumn{4}{c|}{PHEME}               & \multicolumn{4}{c}{HiSim}                 \\
Ratio     & \# Vocab & stance↑ & belief↑ & Ratio     & \# Vocab & stance↑ & content↑ \\ \midrule
0.2       & 31.5K    & 63.80   & 44.16   & 0.2       & 48.2K    & 70.34   & 30.48    \\
0.4       & 31.8K    & 63.13   & 43.57   & 0.4       & 49.3K    & 70.41   & 30.09    \\
0.6       & 32.6K    & 64.10   & 44.25   & 0.6       & 50.9K    & 69.64   & 29.11    \\
0.8       & 34.0K    & 65.73   & 44.65   & 0.8       & 52.8K    & 69.26   & 29.55    \\
Llama-3.1 & 128.3K   & 66.21   & 42.44   & Llama-3.1 & 128.3K   & 70.30   & 30.23    \\ \bottomrule
\end{tabular}
}
\caption{Performance of the simulations when using different vocabularies. \textit{Ratio} represents the reserving ratio for each semantic cluster when filtering words. We at least keep one word for each synonym set.}
\label{tab:exp_vocab}
\end{table}
\begin{table*}[!t]
\resizebox{\textwidth}{!}{
\begin{tabular}{@{}c|cccccc|ccccccc@{}}
\toprule
\multirow{2}{*}{Method} & \multicolumn{6}{c|}{PHEME}                                                  & \multicolumn{7}{c}{HiSim}                                                                    \\
                        & stance↑ & belief↑ & belief\_JS↓ & $token_r$↓ & $token_p$↓ & $token_c$↓ & stance↑ & content↑ & $\Delta bias$ ↓ & $\Delta div$ ↓ & $token_r$↓ & $token_p$↓ & $token_c$↓ \\ \midrule
Qwen                    & 63.35   & 49.25        & 0.1426      & 1.93K      & 78.21K     & 7.36K      & 71.63   & 26.06    & 0.1025          & 0.0246         & 17.68K     & 1.81M      & 214.11K    \\
Qwen w/ Rule            & 62.95   & 51.65        & 0.1475      & 1.81K      & 78.36K     & 7.09K      & 72.04   & 26.77    & 0.0978          & 0.0255         & 14.71K     & 1.77M      & 188.96K    \\
Mistral                 & 62.98   & 52.39        & 0.1529      & 3.10K      & 96.94K     & 11.60K     & 72.02   & 31.78    & 0.1220           & 0.0536         & 26.91K     & 2.36M      & 416.15K    \\
Mistral w/ Rule         & 63.84   & 60.00        & 0.1484      & 2.28K      & 94.76K     & 10.39K     & 72.39   & 32.57    & 0.0963          & 0.0352         & 22.76K     & 2.29M      & 358.23K    \\ \bottomrule
\end{tabular}
}
\caption{Results of simulations driven by Qwen2.5 and Mistral with and without the evolved rule of Llama3.1.}
\label{tab:exp_trans}
\end{table*}

\begin{figure*}[!t]
    \centering
    \includegraphics[width=\linewidth]{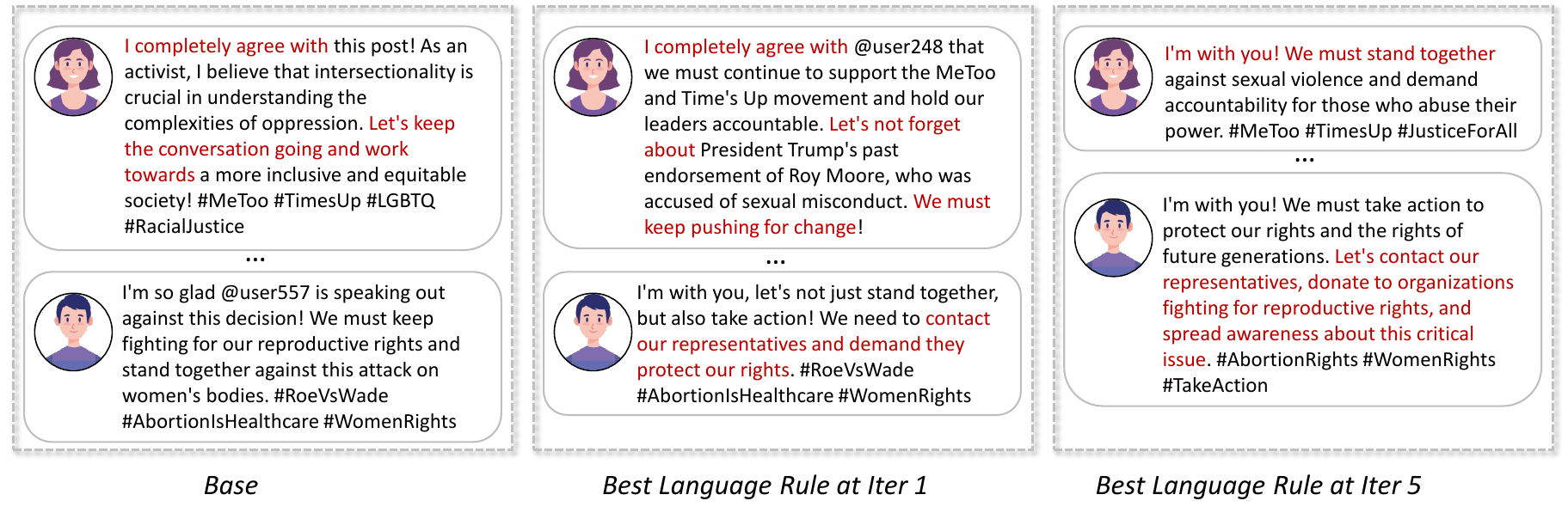}
    \caption{Case study: responses of agents without any communication optimization and with the best evolved rule at iteration 1 and 5. In most cases, agents express more concisely while sometimes fail to follow instructions. }
    \label{fig:case}
\end{figure*}

\subsection{Finer-Grained Evaluation of Languages}
Although the previous part has discussed the effectiveness of different methods in social simulation, where relatively macro dimensions are considered, some may be concerned that language compression could risk losing fine-grained individual semantics. To address this, we conduct a more detailed evaluation of agent language. Table~\ref{tab:exp_fine} reports the semantic and length differences between agent-generated and real user-generated responses across different methods. The metrics include sentence embedding cosine similarity (CosSim), lexical overlap of responses (Jaccard), JS divergence of word distributions (word\_JS), as well as differences in average sentence length ($\Delta_{l_s}$) and word length (($\Delta_{l_w}$) in tokens. We summarize the key findings as follows:

\noindent\textbf{(1) Word Usage Patterns}: \textbf{Even without language compression}, i,e., \textit{Base}, \textbf{agents' responses exhibit low Jaccard similarity with real user tweets}, highlighting the inherent divergence between LLM-generated and human-written texts. This gap is likely due to the agents' limited access to personal or contextual knowledge and the biases introduced by LLMs. Nevertheless, most approaches maintain comparable semantic similarity to \textit{Base}, with \textbf{our method outperforming others} in preserving meaning.

\noindent\textbf{(2) Response Length Patterns}: Agents tend to produce longer and more complex responses than real users, who generally favor brevity and simplicity in social communication. This aligns with the redundancy issues discussed in the introduction section. Compared to baselines, \textbf{our method produces shorter and more concise utterances, which not only improve communication efficiency but also better align with the communication habits of real users in general}.

\subsection{Tracing the Evolution of Language Rules}

To better understand how language rules evolve, we analyze both the progression of dialogue fitness scores and linguistic shifts across iterations, as well as their downstream effects on social simulations.
Figures~\ref{fig:iter}(a) and (b) illustrate the trends observed during the evolution process on synthetic-persona-chat dialogues. Beyond the fitness scores defined in Sec~\ref{sec:rule}, we also track two additional metrics: structural drift and semantic drift~\cite{lazaridou2020multi}. Structural drift assesses fluency and grammaticality relative to natural language, while semantic drift captures how well the generated language preserves the literal meaning of intended targets. Results reveal that as evolution progresses, \textbf{language fitness improves overall: alignment and expressiveness increase, and token consumption decreases}. Simultaneously, both structural and semantic drift decline, \textbf{indicating improved language quality} despite these metrics not being directly optimized during training. \textbf{These improvements translate into downstream gains} as well: as shown in Figure~\ref{fig:iter}(c), simulations guided by the evolved rules demonstrate higher accuracy and lower token usage. However, after several iterations, the fitness score no longer improves, and the optimal rules provided for the simulation tasks remain unchanged, suggesting that the evolution process may have converged.

\subsection{Unpacking the Impact of Vocabulary}
We further explore the impact of the vocabulary on the simulation. As shown in Table~\ref{tab:exp_vocab}, since it is necessary to ensure that at least one word is retained for each semantic cluster, changing the retention ratio has a subtle impact on the size of the vocabulary. Nevertheless, it can be observed that \textbf{the influence of vocabulary size on performance exhibits different trends across simulations}. For PHEME, a larger vocabulary is better, possibly because it covers a more diverse range of topics and discussions, requiring more words for support. In contrast, for HiSim, due to the more focused discussion topics Metoo and Roe, using fewer but more commonly used words can achieve ideal results.


\subsection{Exploring the Transferability of Language Rules Across Different LLMs}
Can the evolved language be used on other models, or do we need to reacquire the language for each model? To answer this question, we apply the acquired language rules to other models, i.e., Qwen2.5-7b-Chat~\cite{qwen2.5} and Mistral-7b-Instruct-v0.3~\cite{jiang2023mistral}. Table~\ref{tab:exp_trans} show that the rules evolved on Llama-3.1 \textbf{can also enable other models to communicate efficiently, again demonstrating the transferability of EcoLANG}.

\subsection{Case Study and Error Analysis}
Figure~\ref{fig:case} showcases some exemplary instances of efficient communication and bad cases. Benefiting from the evolved rule, agents can speak more concisely using words like ``I'm with you'' to replace ``I completely agree with you''. However, sometimes the agents may fail to simplify their expression and disclose excessive details. This may be the result of the model's insufficient ability to follow instructions. A potential solution is to further fine-tune the models using the efficient communication dialogues from the language evolution process. 

\section{Conclusion}
We introduced EcoLANG, a novel two-stage paradigm comprising language evolution and utilization, designed to acquire efficient and effective language for large-scale social simulations. We derive the language by vocabulary compression and rule evolution and demonstrate its applicability across social simulation scenarios. Experiment results highlight EcoLANG's ability to reduce inference costs while maintaining simulation accuracy.

\section*{Limitations}
EcoLANG induces an efficient agent communication language that improves simulation efficiency and reduces inference costs while maintaining simulation accuracy. Despite our careful design, some limitations still exist.
\begin{itemize}
    \item Although EcoLANG improves efficiency, the extent of this improvement is not yet transformative. This is because we focus on reducing token generation but do not address the reduction of the number of inference times. In the future, we plan to integrate it with hybrid frameworks that optimize the number of inference steps, thereby further enhancing efficiency and reducing costs to a greater extent.
    \item Due to the limited available large-scale social simulation datasets for validation, we have currently only tested EcoLANG in PHEME and HiSim, which may raise concerns about its generalizability. In the future, it will be necessary to advance the construction of benchmarks for diverse social simulation scenarios.
    \item Due to the lack of objective and unified evaluation frameworks and metrics for existing LLM-driven social simulations, as compared to task-solving scenarios, we currently partly rely on LLMs to get the fitness value during the selection process, which may introduce potential bias. We will continue to explore more reliable evaluation frameworks for social simulation.

\end{itemize}

\section*{Ethics Statement}
This paper aims to evolve an efficient communication language for social simulation. Like most work in social simulation, it may raise potential considerations and we urge the readers to approach it with caution.
\begin{itemize}
    \item When employing LLMs for social simulation, concerns arise regarding the fidelity and interpretability of the results. If not carefully managed, the risk of bias could exacerbate real-world problems. However, our experiments demonstrate that EcoLANG does not amplify incorrect predictions related to misinformation (PHEME) or opinion polarization (HiSim).
    \item Ensuring the ethical handling of any real-world datasets, including anonymization and consent, is crucial. During our social simulations, all user content was anonymized to minimize privacy risks. 
    \item Although EcoLANG is designed to evolve efficient language, misuse, such as promoting uncivil language, could pose risks. Therefore, strict governance and ethical guidelines should be implemented.
\end{itemize}

\bibliography{acl_latex}

\appendix
\clearpage
\section{Implementation Details of Language Evolution}~\label{app:evo}

\subsection{Vocabulary Compression}
\paragraph{Twitter Corpus for Word Frequency Counting}
Since it's infeasible to get a corpus of all tweets to count words, we have chosen to analyze and gather statistics from existing tweets relevant to the topics of social simulation. Since some tweet links are no longer accessible, we crawled 41,736 tweets from the Twitter 15 and 16 datasets~\cite{liu2015real,ma2016detecting} and 10,673,881 tweets from the social movement dataset~\cite{metoodata,chang2023roeoverturned,blmdata} that were posted before the simulated events in HiSim occurred, resulting in 35,211 and 1,662,657 unique words, respectively.

\paragraph{Semantic Clustering}
We experimented with both top-down clustering, which involves assigning words from the corpus to synsets in WordNet~\cite{miller1995wordnet}, and bottom-up clustering, which encodes each word and groups them into clusters using methods like KMeans or spectral clustering. We found that the top-down approach is more controllable and less likely to group unrelated words into the same cluster, so we adopted the former method.
Specifically, we first remove non-English words, and we compute the center embedding $e_{S_j}$ of each synset $S_j$ in WordNet and calculate the cosine similarity between each candidate word $w_i$ and the center of every synset. The word is then assigned to the synset whose center has the highest similarity, as shown in Eq.~\ref{eq1}.

Due to the fine-grained division of synonym sets in WordNet, many sets contain only one or two words. Therefore, we further merge similar sets using a similarity threshold of 0.8, resulting in 16,545 clusters for PHEME and 47,339 clusters for HiSim.

\paragraph{Intra-Cluster Selection}
Within each semantic cluster, we reserve words with the highest scores calculated by the score function in Eq.~\ref{eq2}. With different reservation ratio $r_w$ for each cluster, we can get vocabularies of different sizes, as shown in Table~\ref{tab:exp_vocab}.

\paragraph{Tokenization}
To ensure normal generation by LLMs, in addition to retaining tokens corresponding to the selected words, we also preserve tokens for the LLM's special tokens, punctuation, abbreviations, and emojis.

\subsection{Rule Evolution}

\subsubsection{Initialization}~\label{app:evo_init}
We initialize the language rules by human crafting and LLM generation. We calculate the information density of each tweet in the Twitter corpus, and summarize rules that can reflect the characteristics of these tweets. For LLMs, we ask GPT-4o how to issue rule instructions to enable efficient communication. Specifically, we obtained the following rule prompts:

\begin{tcolorbox}[width=\linewidth,title={Initial Rules for Evolution}]

1. Please respond concisely.

2. Provide a brief summary of your response.

3. Feel free to replace lengthy words or phrases with hashtags and symbols, like emojis.

4. Please use simple sentence structures.

5. Please omit unnecessary components such as subjects or predicate verbs.

6. Try using abbreviations or slang to shorten your sentences.

7. Identify your main point and communicate it directly without unnecessary details.

8. Avoid repeating ideas and removing unnecessary filler words.

9. Get to the point quickly and clearly, without over-explaining.

10. Remove words like "very" or "really" that don’t add value.

\end{tcolorbox}

\subsubsection{Communication}
We use the validation set of the synthetic-persona-chat dataset for communication simulation. We append the sampled language rule behind the profile of agents in their system prompts. In practice, we use AutoGen~\cite{wu2023autogen} to generate dialogues between agents, and the system prompt used is as follows.

\begin{tcolorbox}[width=\linewidth,title={Prompt of Agents for Communication}]
You are \{agent\_name\}. \{agent\_profile\}

\{few-shot chat history for initialization\}

What will you, \{agent\_name\}, speak next?

\{rule\}
\end{tcolorbox}

\subsubsection{Selection}
For the fitness function in selection value, we set the hyperparameters $\lambda_{align}=1$, $\lambda_{eff}=0.6$ and $\lambda_{exp}=0.6$, learning from previous work~\cite{chen2024optima}. We use the following prompts to instruct GPT-4o to give the alignment score and expressiveness score to the dialogues. 

\begin{tcolorbox}[width=\linewidth,title={Prompt for Alignment Evaluation}]
Please evaluate whether the agents' responses align with the persona reflected in the reference response. 

Please focus on the aspects of content, emotion and atttude, and ignore differences in language structure, e.g., word choice, sentence length, emoji usage and syntax.

Agent's response: \{simulated\_dialog\}

Reference response: \{reference\_dialog\}

Please rate on a scale of 1 to 5, with 1 being most inconsistent and 5 being most like the persona.

Please write a short reason and strictly follow the JSON format for your response:

\{\{"reason": <str>, "score": <int>\}\}

\end{tcolorbox}

\begin{tcolorbox}[width=\linewidth,title={Prompt for Expressiveness Evaluation}]
Please evaluate whether the agents' language is clear and easy to understand.

Agents' language: \{simulated\_dialog\}

Please rate on a scale of 1 to 5, with 1 being most unclear and 5 being most clear.

Please write a short reason and strictly follow the JSON format for your response:

\{\{"reason": <str>, "score": <int>\}\}

\end{tcolorbox}

\begin{figure}[!t]
    \centering
    \includegraphics[width=\linewidth]{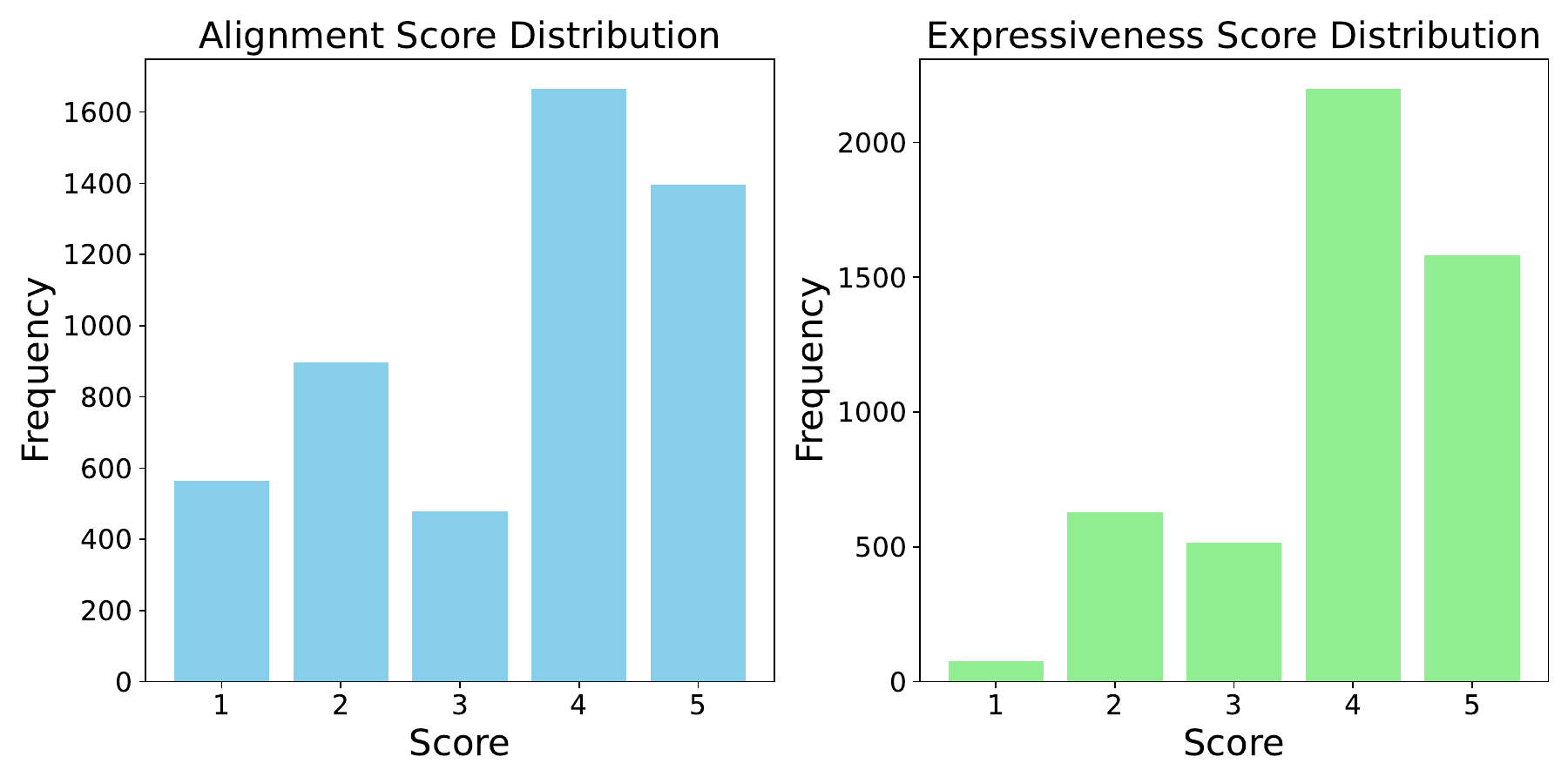}
    \caption{Alignment and expressiveness score distribution in the first iteration.}
    \label{fig:score_dist}
\end{figure}

Figure~\ref{fig:score_dist} shows the score distribution of dialogues in iteration 1, indicating that the judge model GPT-4o is capable of assigning differentiated scores. In addition, we sampled 50 dialogues for human annotation and found that GPT-4o is more consistent (Cohen's Kappa: 0.48) with human judgments than GPT-4o-mini. Therefore, we chose GPT-4o as the judge model.

\subsubsection{Crossover \& Mutation}
We use the following prompts to conduct crossover and mutation on parent rules.

\begin{tcolorbox}[width=\linewidth,title={Prompt for Crossover}]
Please cross over the following prompts and generate a new prompt bracketed with <prompt> and </prompt>.

Prompt 1: \{rule\_prompt1\}

Prompt 2: \{rule\_prompt2\}
\end{tcolorbox}

\begin{tcolorbox}[width=\linewidth,title={Prompt for Mutation}]
Mutate the prompt and generate a new prompt bracketed with <prompt> and </prompt>

Prompt: \{rule\_prompt\}

\end{tcolorbox}

\subsubsection{Update and Iteration}
In each iteration, we adopt the elitism strategy of genetic algorithm to reserve the top-5 rules in current population and generate 5 new rules through crossover and mutation. The overall process for the evolution can be described in Algorithm~\ref{alg:evo}.

\begin{algorithm}
\caption{Evolution of the language rules}
\begin{algorithmic}[1]
    \REQUIRE Initial rules $\mathcal{P}_1=\left\{p_1, p_2, \ldots, p_N\right\}$, size of rule population $N$, a set of scenarios for dialogue simulation $\mathcal{D}=\{d_i\}$, number of sampled rules for each scenario $M$, a pre-defined number of iterations $T$, fitness function for each dialogue $F$, crossover and mutation operation $Opr(\cdot)$, update strategy $Upd(\cdot)$
    \FOR{$t$ in $1$ to $T$}
        \STATE \textbf{Communication}: sample and assign rules to each scenario $d_i$ and use LLM-driven agents to generate dialogues $\left\{\tau_i^j\right\}_{j=1}^M$ in these scenarios
        \STATE \textbf{Selection}: use the fitness function to evaluate the dialogues $s_i^j \leftarrow F(\tau_i^j)$, and average the scores of the dialogues based on rules used to get fitness of each rule
        \STATE \textbf{Crossover and Mutation}: select a certain number of rules as parent rules $p_{r_1}, \ldots, p_{r_k} \sim \mathcal{P}_t$, and generate new rules based on the parent rules by leveraging LLMs to perform crossover and mutation $\{p_i^{\prime}\} \leftarrow {Opr}\left(p_{r_1}, \ldots, p_{r_k}\right)$
        \STATE \textbf{Update}: update the set of rules $\mathcal{P}_{t+1} \leftarrow Upd(\mathcal{P}_t, \{p_i^{\prime}\})$
        
    \ENDFOR
    
    \RETURN the best rule $p^*_t$ at each iteration $t$ 
\end{algorithmic}
\label{alg:evo}
\end{algorithm}

\subsubsection{Evolved Rules}
Based on the vocabularies of PHEME and HiSim, we perform rule evolution using the synthetic-persona-chat dataset. In each iteration, we obtain the following best rules:

\begin{tcolorbox}[width=\linewidth,title={Best Rules for PHEME}]
iter 1: Please use simple sentence structures.

iter 2: Respond briefly, removing unnecessary words.

iter 3: Eliminate repetitive ideas, unnecessary fillers, and respond concisely.

iter 4: Eliminate repetitive ideas, unnecessary fillers, and respond concisely.

iter 5: Remove redundancy, filler words, and respond briefly.

\end{tcolorbox}

\begin{tcolorbox}[width=\linewidth,title={Best Rules for HiSim}]
iter 1: Avoid repeating ideas and removing unnecessary filler words.

iter 2: Please use simple sentence structures.

iter 3: Eliminate redundancy, cut filler, and be concise.

iter 4: Eliminate redundancy, cut filler, and be concise.

iter 5: Eliminate redundancy, cut filler, and be concise.

\end{tcolorbox}
\section{Implementation Details of Language Utilization (Social Simulation)}~\label{app:simu}

\subsection{Implementation Details}
All the simulations are conducted in OASIS framework~\cite{yang2024oasis}. We run the simulator on a Linux
server with 8 NVIDIA GeForce RTX 4090 24GB GPU and an Intel(R) Xeon(R) Gold 6226R CPU. We run each simulation three times and report the average results to reduce randomness.

\begin{table}[]
\resizebox{0.48\textwidth}{!}{
\begin{tabular}{@{}cc@{}}
\toprule
\textbf{Hyperparameter} & \textbf{Value}        \\ \midrule
model                   & Llama-3.1-8B-Instruct \\
temperature             & 0                     \\
max\_tokens              & 512                   \\
num\_steps               & max depth of each (non)rumor                   \\ \bottomrule
\end{tabular}
}
\caption{Hyperparameters of PHEME Simulation.}
\label{tab:app-pheme}
\end{table}

\subsection{PHEME Simulation}\label{app:pheme_smiu}
We initialize the agents with user profiles and network information acquired from the PHEME dataset. We prompt GPT-4o-mini to write a short description given each user's biography on Twitter. For each instance in PHEME, we only retain replies with content for simulation and validation. The action space prompt for PHEME in OASIS simulation is as follows and the hyperparameters are shown in Table~\ref{tab:app-pheme}. Other parameters and mechanisms, such as the memory mechanism, are set to the defaults in the OASIS framework.

\begin{tcolorbox}[width=\linewidth,title={Action Space Prompt for PHEME in OASIS}]
You're a Twitter user, and I'll present you with some posts. After you see the posts, choose some actions from the following functions.

Suppose you are a real Twitter user. Please simulate real behavior.
\\

- do\_nothing: Most of the time, you just don't feel like reposting or liking a post, and you just want to look at it. In such cases, choose this action "do\_nothing"

- quote\_post: Quote a specified post with given content.

\ \ \- Arguments:
    
\ \ \ \ \ \    - "post\_id" (integer) - The ID of the post to be quoted.
        
\ \ \ \ \ \    - "quote\_content" (string) - The content of the quote. You can `quote\_post’ when you want to share a post while adding your own thoughts or context to it.

\{rule\_prompt\}
\end{tcolorbox}

\subsection{PHEME Evaluation}
For simulation results on PHEME, we include the following metrics to evaluate simulation effectiveness:
\begin{itemize}
    \item Stance Consistency: we label the stance of each agent's and real user's \textit{initial} response towards the source tweet given the tree-like threads, with the label space being \textit{support}, \textit{deny}, \textit{query} and \textit{comment} from~\cite{derczynski2017semeval}. 
    \item Belief Consistency: Following ~\cite{liufps}, we label the belief of agents and real users at the \textit{end} of simulation. Since we observed that some agents or users did not explicitly express belief or disbelief, we added an additional category, \textit{unknown}, alongside the existing \textit{belief} and \textit{disbelief} labels.
    \item Belief JS Divergence: To measure the belief distribution of the user group regarding fake news, we additionally incorporated the JS divergence of the belief distribution to assess the effectiveness of the simulation at the group level.
\end{itemize}

The prompts for stance and belief annotation are as follows.

\begin{tcolorbox}[width=\linewidth,title={Prompt for PHEME Stance Labeling}]
Given threads discussing a news, please label the stance of the question tweet on the source news tweet.
\\

Treads: \{threads\}

Question tweet: \{tweet\}
\\

Please choose from the following options:

1. support: the author of the response supports the veracity of the news.

2. deny: the author of the response denies the veracity of the news.

3. query: the author of the response asks for additional evidence in relation to the veracity of the news.

4. comment: the author of the response makes their own comment without a clear contribution to assessing the veracity of the news.
\\

Please strictly follow the JSON format for your response:

\{\{"stance": <str>\}\}

\end{tcolorbox}

\begin{tcolorbox}[width=\linewidth,title={Prompt for PHEME Belief Labeling}]
Please determine whether the author of the final tweet believes the source news.
\\

Source News:\{source\_tweet\}

Final Tweet:\{final\_tweet\}
\\

If the author does not believe the source news, questions the AUTHENTICITY of the source news or queries for more information about the AUTHENTICITY of the news, please label it as disbelief.

If the author expresses opinions or call for actions under the assumption that the news is true, please label it as belief.

If the author discusses something unrelated to the source news, please label it as unknown. Please label 0 for disbelief, 1 for belief and 2 for unknown.
\\

Please write a short reason and strictly follow the JSON format for your response:

\{\{"reason": <str>, "label": <int>\}\}

\end{tcolorbox}

\subsection{HiSim Simulation}\label{app:hisim_simu}
Metoo and Roe datasets in HiSim provide profiles and historical tweets of 1,000 users respectively, as well as their social networks in Twitter. We use this information to initialize the agents in the OASIS platform. To reduce the randomness introduced by the OASIS platform, we ban the recommendation systems and only enable agents to get information from external news and who they are following. The action space prompt for PHEME in OASIS simulation is as follows. The hyperparameters are shown in Table~\ref{tab:app-hisim}. Other parameters and mechanisms, such as the memory mechanism, are set to the defaults in the OASIS framework.

\begin{tcolorbox}[width=\linewidth,title={Action Space Prompt for HiSim in OASIS}]
You're a Twitter user, and I'll present you with some posts. After you see the posts, choose some actions from the following functions.

Suppose you are a real Twitter user. Please simulate real behavior.
\\

- do\_nothing: Most of the time, you just don't feel like reposting or liking a post, and you just want to look at it. In such cases, choose this action "do\_nothing"

- create\_post: Create a new post with the given content.

\ \ \    - Arguments: "content" (str): The content of the post to be created.

- repost: Repost a post.

\ \ \     - Arguments: "post\_id" (integer) - The ID of the post to be reposted. You can `repost` when you want to spread it.

- quote\_post: Quote a specified post with given content.

\ \ \  - Arguments:

\ \ \ \ \ \        - "post\_id" (integer) - The ID of the post to be quoted.

\ \ \ \ \ \        - "quote\_content" (string) - The content of the quote. You can `quote\_post‘ when you want to share a post while adding your own thoughts or context to it.

\{rule\_prompt\}
\end{tcolorbox}

\begin{table}[]
\centering
\begin{tabular}{@{}cc@{}}
\toprule
\textbf{Hyperparameter} & \textbf{Value}        \\ \midrule
model                   & Llama-3.1-8B-Instruct \\
temperature             & 0                     \\
max\_tokens              & 512                   \\
num\_steps               & 14                    \\ \bottomrule
\end{tabular}

\caption{Hyperparameters of HiSim Simulation.}
\label{tab:app-hisim}
\end{table}

\subsection{HiSim Evaluation}
For simulation results on HiSim, we follow~\cite{mou-etal-2024-unveiling} to include the following metrics to evaluate simulation effectiveness:
\begin{itemize}
    \item Stance Consistency: we classify the \textit{initial} response of agents and real users into three categories: \textit{support}, \textit{neutral} and \textit{oppose}, towards the given target \textit{\#Metoo movement} and \textit{the protection of abortion rights}, and compute the consistency between agents and users.
    \item Content Consistency: we classify the \textit{initial} response of agents and real users into 5 types, i.e., \textit{call for action}, \textit{sharing of opinion}, \textit{reference to a third party}, \textit{testimony} and \textit{other}.
    \item $\Delta bias$ and $\Delta div$: bias is measured as the deviation of the mean attitude from the neutral attitude, while diversity is quantified as the standard deviation of attitudes. These metrics are calculated at each time step and averaged over time. The differences between the simulated and real-world measures, denoted as $\Delta bias$ and $\Delta div$ are reported.
\end{itemize}

The prompts for stance and content labeling are borrowed from~\cite{mou-etal-2024-unveiling}. Notably, we focus on the macro setting from the original HiSim paper, which involves continuous, multi-turn interactions to simulate complex social dynamics over time. However, we did not include HiSim as a baseline, as it adopts a different agent architecture based on AgentVerse from our implementation on OASIS.

\subsection{Evaluation Bias}

\begin{table}[]
\centering
\begin{tabular}{@{}cc@{}}
\toprule
\textbf{Dim.} & \textbf{Consistency}        \\ \midrule
stance             & 0.94                     \\
belief              & 0.78                  \\\bottomrule
\end{tabular}
\caption{Consistency of GPT-4o-mini judging the stance and belief when taking human evaluations as the ground-truth reference.
}
\label{tab:app-human}
\end{table}

\begin{table*}[!t]
\centering
\resizebox{0.9\textwidth}{!}{
\begin{tabular}{@{}l|cccccccccc@{}}
\toprule
\multicolumn{1}{c|}{\textbf{Method}} & \textbf{stance↑} & \textbf{content↑} & \textbf{$\Delta{bias}$↓} & \textbf{$\Delta{div}$↓} & \textbf{$token_r$↓} & \textbf{$token_p$↓} & \textbf{$token_c$↓} & \textbf{CosSim↑} & \textbf{Jaccard↑} & \textbf{word\_JS↑} \\ \midrule
AgentTorch                           & 67.87            & 31.81             & 0.098                    & 0.024                   & 2.5K                & 0.48M               & 94.34K              & 0.666              & 0.058               & 0.360              \\ \midrule
w/ Summary                           & 68.28            & 32.27             & 0.151                    & 0.032                   & 1.5K                & 0.48M               & 88.58K              & 0.698              & 0.054               & 0.366              \\
w/ AutoForm                          & 65.19            & 32.52             & 0.092                    & 0.026                   & 1.8K                & 0.47M               & 84.09K              & 0.726              & 0.065               & 0.359              \\
w/ KQML                              & 67.64            & 32.20             & 0.110                    & 0.018                   & 1.5K                & 0.49M               & 88.90K              & 0.693              & 0.064               & 0.361              \\ \midrule
w/ Vocab                             & 67.61            & \textbf{33.56}    & 0.098                    & \textbf{0.013}          & 1.6K                & 0.47M               & 91.39K              & 0.726              & \textbf{0.066}      & 0.359              \\
w/ Rule                              & 67.76            & 33.35             & \textbf{0.086}           & 0.015                   & 1.3K                & 0.47M               & 80.96K              & 0.735              & 0.066               & \textbf{0.356}     \\
w/ EcoLANG                           & \textbf{68.63}   & 33.45             & 0.099                    & 0.017                   & \textbf{1.2K}       & \textbf{0.46M}      & \textbf{78.03K}     & \textbf{0.740}     & 0.066               & 0.358              \\ \bottomrule
\end{tabular}
}
\caption{The results of the communication simplification method combined with AgentTorch on HiSim dataset. Only 36 prototype agents were used in all experiments. The number of prototypes is determined by the combination of fundamental attributes such as gender, political inclination, and activity level.}
\label{tab:exp_agenttorch}
\end{table*}

Since we partially rely on LLMs for evaluation, this approach may introduce some evaluation bias. To address this, we sample 100 simulation instances and instruct two human annotators to label the stance and belief of the responses, providing them with the same information as given to GPT. Table~\ref{tab:app-human} shows the consistency between the annotations of GPT-4o-mini and those of the human annotators.

\subsection{Integration with AgentTorch}\label{app:agenttorch}
The communication simplification methods are orthogonal to AgentTorch~\cite{chopra2024limits} and can be combined with it to enhance efficiency further. To understand the potential of combining different communication simplification methods with this paradigm, we conducted experiments by integrating different communication simplification methods with AgentTorch. Since each scenario in PHEME involves a relatively small number of agents, further clustering them into a few prototypes would overly simplify the agent population, resulting in homogeneous content and limiting the generation of meaningful responses. Given these limitations, we determined that AgentTorch is not a suitable baseline for PHEME and therefore conducted experiments only on HiSim.

The results in Table~\ref{tab:exp_agenttorch} show that combining all methods with AgentTorch can further improve simulation efficiency, reducing token consumption by up to an additional 80\% compared to Table~\ref{tab:exp_main}. Among these, our method demonstrates advantages in both effectiveness and efficiency, highlighting its robustness. However, integrating with AgentTorch has some side effects. While using a small number of agents drastically reduces token usage, it also compromises the diversity and accuracy of agent responses, leading to noticeable shortcomings in content-related metrics, e.g., stance and CosSim, compared to results in Table~\ref{tab:exp_main} and Table~\ref{tab:exp_fine}.

\end{document}